%% file: PESGA_2027_paper.tex
\documentclass[conference]{IEEEtran}
\IEEEoverridecommandlockouts

\usepackage{cite}
\usepackage{amsmath,amssymb,amsfonts}
\usepackage{algorithmic}
\usepackage{graphicx}
\usepackage{booktabs}
\usepackage{textcomp}
\usepackage{xcolor}
\usepackage{pgfplots}
\pgfplotsset{compat=1.18}
\usetikzlibrary{arrows.meta,calc}
\def\BibTeX{{\rm B\kern-.05em{\sc i\kern-.025em b}\kern-.08em
    T\kern-.1667em\lower.7ex\hbox{E}\kern-.125emX}}

\usepackage{eso-pic}
\AddToShipoutPictureFG{%
  \AtPageLowerLeft{%
    \makebox[\paperwidth][c]{\raisebox{6mm}{\footnotesize\itshape This preprint is currently under peer review.}}}}

\begin{document}
\bstctlcite{IEEEexample:BSTcontrol}

\title{Which Model Belongs in the Loop? Evaluating Fitness Functions for Network Topology Optimization with a Quality-Diversity Algorithm\\
}

\author{
\IEEEauthorblockN{Paul Bannmüller\IEEEauthorrefmark{1}, Olayiwola Arowolo\IEEEauthorrefmark{1}, Jochen Stiasny\IEEEauthorrefmark{1}, Jochen Cremer\IEEEauthorrefmark{1} \IEEEauthorrefmark{7}}
\IEEEauthorblockA{\IEEEauthorrefmark{1} Department of Electrical Sustainable Energy, Delft University of Technology, Delft, The Netherlands\\
\IEEEauthorrefmark{7} AIT Austrian Institute of Technology GmbH, Vienna, Austria\\
p.e.bannmuller@tudelft.nl}

}

\maketitle

\begin{abstract}
Network topology optimization (NTO) can reduce redispatch cost by rerouting power flows in transmission systems.
To approach NTO, a complex combinatorial problem, gradient-free optimization algorithms that leverage diversity have shown promising results.
These algorithms rely on frequent evaluations of candidate topologies by fitness functions. Recent work has shown that fast DC power flow-based fitness functions can serve this purpose.
However, DC power flow fitness functions can misjudge the cost and feasibility of topologies, potentially leading to wrong operational decisions.
Understanding the impact of fitness function accuracy on optimization performance is therefore crucial.
This study compares the optimization performance of four fitness functions based on DC and AC power flow and on DC and AC optimal power flow.
Furthermore, we propose a ranking-based metric to estimate the optimization performance of a fitness function on a curated dataset.
Case studies cover the IEEE 118-bus case with 50 congested operating points and a 793-bus case for validation.
Results show that optimal power flow-based fitness functions perform better in the optimization loop than power flow-based ones.
Additionally, the results indicate that the metric can predict the optimization performance order of the fitness functions, but only if the distribution shift between the curated dataset and the search path of the optimization is small.
\end{abstract}


\section{Introduction}\label{sec:intro}

\begin{figure}[!t]
\centering
\input{figures/surrogate_schematic.tex}
\caption{Schematic failure modes of fitness functions over the topology space, adapted from~\cite{jinSurrogateassistedEvolutionaryComputation2011}. The tight fitness function excludes the global optimum, the loose one ranks a redispatch-infeasible topology best, and the good one ranks topologies like the redispatch cost despite its offset.}
\label{fig:failure_modes}
\end{figure}
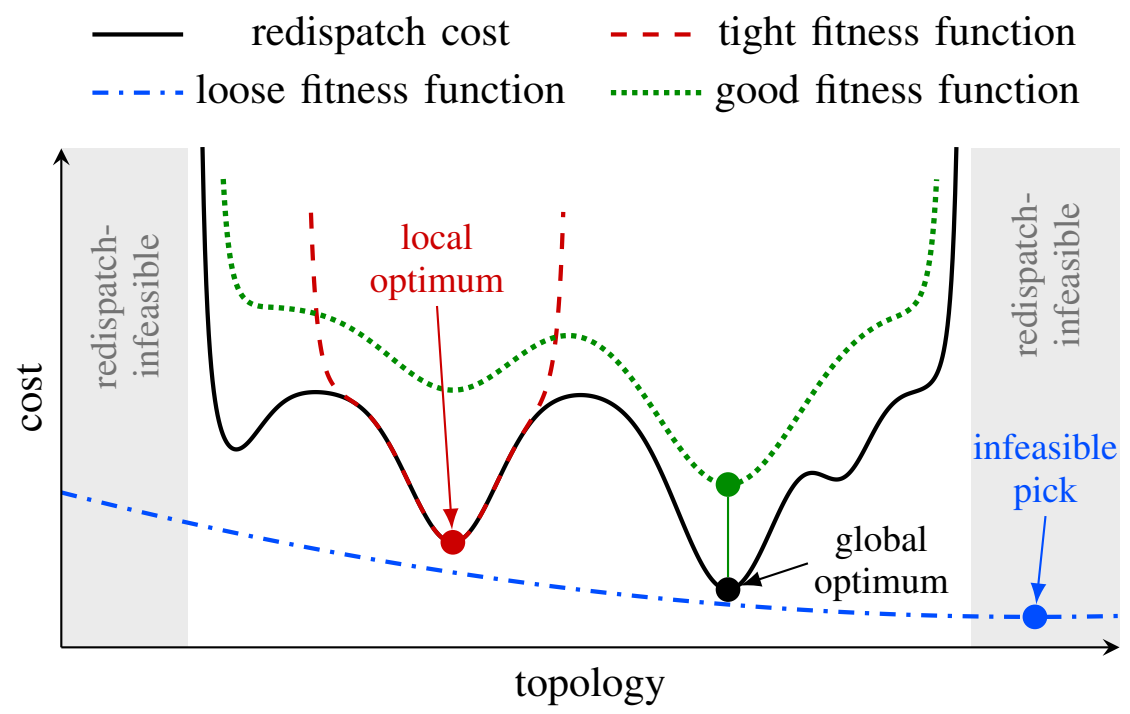

Increasing intermittent energy generation, electrification, and large loads increase congestion in the transmission system.
In the EU, transmission system operators (TSOs) tackle congestion with remedial actions, often by redispatching generators.
This redispatch increased electricity costs by 5.4 billion euros in 2025~\cite{acerProgressEUElectricityWholesale2026}.
Network topology optimization (NTO) aims to reduce the congestion cost by rerouting power flows through switching lines, splitting busbars, or reconfiguring substations~\cite{heidarifarNetworkTopologyOptimization2016}.

Finding the optimal topology is a complex combinatorial problem that can be modeled as a mixed-integer program~\cite{heidarifarNetworkTopologyOptimization2016}.
However, the power flow equations introduce non-convex constraints, resulting in a mixed-integer nonlinear program (MINLP) that cannot be solved at scale.
Applying linear or convex (e.g., second-order cone) power flow approximations accelerates the optimization, but may lead to suboptimal or infeasible solutions~\cite{bakerSolutionsDCOPF2020}.
Applying learning-based approaches often bases on labeled data which requires solving the MINLP~\cite{martinezSelfSupervisedGraphNeural2026}.
Approaches that need no labeled data like unsupervised learning still require simplifying and solving the MINLP for training~\cite{martinezSelfSupervisedGraphNeural2026} or, in reinforcement learning, struggle with the large combinatorial action space and rely heavily on heuristics to reduce it~\cite{sarOptimizingPowerGrid2025}.

Gradient-free optimization algorithms over the discrete topology space are often used for action space reduction~\cite{preuschoffOptimizingSwitchingActions2025,sarOptimizingPowerGrid2025}, but also to find solutions directly~\cite{lytaevExtendedGeneticAlgorithms2026,westerbeckTransmissionTopologyOptimization2026,groeneveldExactEvolutionaryAlgorithms2026}.
These algorithms rely on frequent evaluations of candidate topologies by fitness functions.
The German TSO 50Hertz and the Belgian TSO Elia recently published ToOp~\cite{westerbeckTransmissionTopologyOptimization2026}, an NTO tool that they test in their grid operation.
ToOp leverages the quality-diversity algorithm MAP-Elites, a genetic algorithm that maintains an archive of diverse solutions called elites, in combination with a highly parallelized DC power flow (DCPF) with low-rank updates based on distribution factors~\cite{dijkUnifiedAlgebraicDeviation2024}.
ToOp integrates into the current workflow of the TSOs, where they optimize the topology based on the economic dispatch and then optimize redispatch in a sequential step.

This workflow setting has two shortcomings.
The first is the mismatch of objectives: minimizing DC overload does not necessarily yield topologies that reduce redispatch cost compared to the base topology or are AC-feasible~\cite{lytaevExtendedGeneticAlgorithms2026, westerbeckConstraintsCongestionbasedTransmission2026}.
Quality-diversity optimization partly mitigates this, because a diverse set of elites can absorb model inaccuracies, but it gives no guarantees~\cite{bannmullerAddressingModelInaccuracies2025,crozierModelingGenerateAlternatives2025a}.
The second is the decoupled optimization of topology and redispatch, which leads to suboptimal solutions~\cite{ewerszumrodeIterativeApproachGrid2024}.

Current research aim at faster AC formulations and learned proxies that solve power flow and optimal power flow tasks more quickly, reducing model inaccuracies in the optimization loop.
Examples are GPU-based ACPF~\cite{lytaevExtendedGeneticAlgorithms2026,hunkemollerGPUbasedNewtonRaphson2026} and machine-learned AC optimal power flow (ACOPF) proxies~\cite{arowoloGeneralizationGraphNeural2025,puechGENCOUnifiedNeural2026}.

The literature on surrogate-assisted optimization algorithms shows that comparison-based optimizers need surrogates that rank well, not necessarily surrogates with low approximation error~\cite{jinQualityMeasuresApproximate,jinSurrogateassistedEvolutionaryComputation2011,loshchilovComparisonBasedOptimizersNeed2010}.
This paradigm can be translated to fitness functions in NTO: the fitness function does not have to be accurate, but it has to rank topologies in a way that is consistent with the true redispatch cost, as illustrated in Fig.~\ref{fig:failure_modes}.
Yet, to the authors' knowledge, no metric has been proposed that can be used in the development of fitness functions to estimate the optimization performance for NTO.

This paper makes two contributions:
\begin{itemize}
\item We compare the optimization performance of four fitness functions based on DCPF, ACPF, DCOPF, and ACOPF with ToOp's MAP-Elites, and observe the impact of the diverse set of elites on the optimization performance.
\item We propose a ranking metric to estimate the optimization performance of a fitness function and observe the problem of a distribution shift between the dataset the metric is computed on and the search path of the optimization.
This metric can help future researchers and practitioners choose and develop appropriate fitness functions for the problem.
\end{itemize}

\section{Methodology}\label{sec:method}

\subsection{NTO problem formulation}

In this study, we formulate the NTO problem as finding a topology that minimizes the redispatch cost.
For a topology $z$, let $g_m(x,u;z)=0$, $m\in\{\mathrm{AC},\mathrm{DC}\}$, be the power flow equations of pandapower~\cite{thurnerPandapowerOpenSourcePython2018}, with the dispatch $u=P_g$.
In AC, $g_{\mathrm{AC}}$ is the nodal active and reactive power balance with state $x=(\theta,V,Q_g)$; $g_{\mathrm{DC}}$ is its lossless linear approximation with $V=1$\,pu and $x=\theta$.
We write $\phi_\ell$ for the flow on line $\ell$: $S_\ell$ in AC and $P_\ell$ in DC.
The OPF on $z$ is
\begin{equation}
C_m(z)=\min_{x,u}\textstyle\sum_g c_g(P_g)\ \text{s.t.}\ g_m(x,u;z)=0,\,h_m(\cdot)\le 0,
\label{eq:opf}
\end{equation}
where $h_m$ holds the line limits $|\phi_\ell|\le\overline{S}_\ell$, the generator limits on $P_g$ and, in AC, the limits on $V$ and $Q_g$.
The zonal market clears with the economic dispatch $u^{\mathrm{e}}$, the merit-order dispatch that respects only the system-wide power balance and the generator limits, as in European zonal markets such as the Netherlands.
Its cost $C^{\mathrm{e}}$ does not depend on $z$.
The TSO then redispatches the generators to a dispatch that is feasible on $z$, which we model as the ACOPF, so the redispatch cost is $R(z)=C_{\mathrm{AC}}(z)-C^{\mathrm{e}}$.
We evaluate $z$ against the base topology $z_0$, where $\Delta C(z)<0$ is a redispatch saving:
\begin{equation}
\Delta C(z)=\frac{R(z)-R(z_0)}{C_{\mathrm{AC}}(z_0)}=\frac{C_{\mathrm{AC}}(z)-C_{\mathrm{AC}}(z_0)}{C_{\mathrm{AC}}(z_0)}.
\label{eq:deltaC}
\end{equation}
This assumes that redispatch is priced at the generator cost functions in both directions and that every ACOPF dispatch is reachable from $u^{\mathrm{e}}$, i.e., it neglects ramping and the tighter redispatch limits around $u^{\mathrm{e}}$.

\subsection{MAP-Elites for NTO}\label{sec:mapelites}

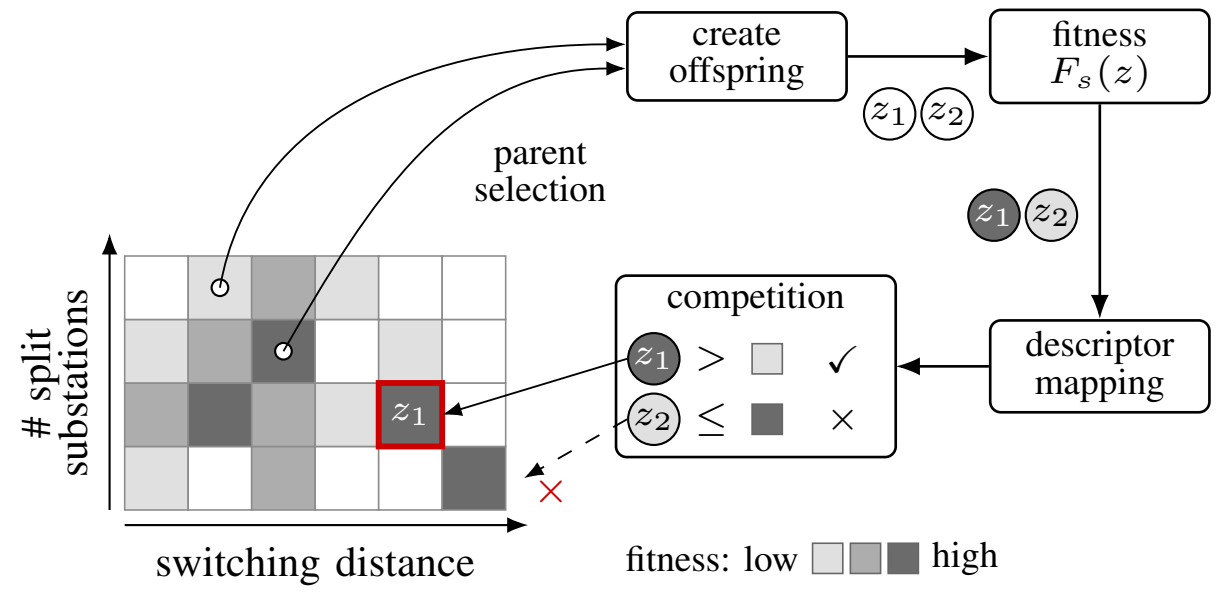
\begin{figure}[!t]
\centering
\input{figures/mapelites_schematic.tex}
\caption{One generation of MAP-Elites in ToOp. Parents are drawn uniformly from the occupied cells. Each offspring is evaluated with $F_s$~\eqref{eq:fitness} and assigned to a cell by its two behavior descriptors. It replaces the elite of its cell only if its fitness is strictly higher ($z_1$), and is discarded otherwise ($z_2$); empty cells are always filled. Darker cells hold elites of higher fitness.}
\label{fig:mapelites}
\end{figure}

We use the gradient-free, quality-diversity optimization algorithm MAP-Elites in its implementation of ToOp~\cite{westerbeckTransmissionTopologyOptimization2026}.
The genetic algorithm iteratively creates candidate topologies over a number of generations, evaluates them with a fitness function that it aims to maximize, and keeps a diverse set of topologies in an archive, as shown in Fig.~\ref{fig:mapelites}.

\subsection{Fitness functions}\label{sec:fitness}

This study compares four fitness functions: those based on DCPF, ACPF, and DCOPF approximate the ACOPF fitness, which serves as the reference.
PF-based fitness functions fix $u=u^{\mathrm{e}}$, solve $g_m=0$ for $x$, and use the negative sum of the line overloads, as is the default in ToOp~\cite{westerbeckTransmissionTopologyOptimization2026}.
OPF-based fitness functions solve \eqref{eq:opf} and use the negative cost normalized by the base topology.
With $m$, the model of fitness function $s$, MAP-Elites maximizes
\begin{equation}
F_s(z)=
\begin{cases}
-10^{9} & \text{islanded/no solution,}\\
-\sum_\ell \max(0,|\phi_\ell|-\overline{S}_\ell) & \text{$s$ PF-based,}\\
-C_m(z)/C_m(z_0) & \text{$s$ OPF-based.}
\end{cases}
\label{eq:fitness}
\end{equation}
Islanding is checked on the topology itself, so islanded topologies are removed and penalized before the model evaluates them.
\emph{No solution} means a non-converged ACPF or a non-converged or infeasible DCOPF or ACOPF; a topology is \emph{ACOPF-solvable} if its ACOPF has a solution.
The penalty lies far below any real fitness, so a solvable topology always beats an unsolvable one.

We consider the N-0 state only, without N-1 contingencies.
A single slack bus, held at $V=1$\,pu by the pandapower OPF, covers all losses without limits on its active power.
The ACPF holds the generator voltages at the setpoints of the case file and enforces the generator reactive power limits, with one retry from a DC power flow start if it fails, whereas the ACOPF optimizes the generator voltages within the bus voltage limits.

\subsection{Ranking metric to predict optimization performance}\label{sec:metric}
Fitness functions for NTO should model both the cost and the feasibility of topologies in a way that enables consistent ranking, as illustrated in Fig.~\ref{fig:failure_modes}.
We propose a metric that measures a fitness function's capability of ranking topologies considering redispatch cost and feasibility: the Kendall $\tau$\footnote[1]{We use its $\tau_b$ variant~\cite{kendallTreatmentTiesRanking1945}, which accounts for ties (all penalized topologies tie) and is more robust on small sets than Spearman's $\rho$. We write $\tau$ for $\tau_b$.
} \cite{jinQualityMeasuresApproximate,lyuLearnabilityOfflineModelBased2026}. %

For a set $\mathcal{D}$ of $n$ topologies, a pair $(z, z')$ is concordant if the penalized fitness $F_s$~\eqref{eq:fitness} and the reference fitness $F^{\star} = F_{\mathrm{ACOPF}}$ order it the same way, and discordant if they order it oppositely:
\begin{equation}
\tau(\mathcal{D}) = \frac{n_{\mathrm{con}} - n_{\mathrm{dis}}}{\sqrt{(n_0 - n_{s})(n_0 - n_{\star})}} ,
\label{eq:taub}
\end{equation}
with the numbers of concordant and discordant pairs $n_{\mathrm{con}}$ and $n_{\mathrm{dis}}$, all pairs $n_0 = n(n-1)/2$, and the numbers of pairs tied in $F_s$ and in $F^{\star}$, $n_{s}$ and $n_{\star}$.
$\tau$ lies in $[-1, 1]$; it is 1 if both fitness functions rank $\mathcal{D}$ identically, and 0 if the ranking is right in 50\,\% of the cases.

To adapt the metric to the MAP-Elites algorithm, we compute $\tau$ \emph{per cell}, because MAP-Elites only compares candidates within the same cell of the archive.
With $\mathcal{D}_k$ the topologies of a dataset that fall into cell $k$, we aggregate by the median over cells,
\begin{equation}
\bar{\tau} = \operatorname*{median}_{k \in \mathcal{C}} \, \tau(\mathcal{D}_k) ,
\label{eq:taucell}
\end{equation}
where $\mathcal{C}$ holds the cells with at least 8 solved samples, because $\tau$ is noisy for smaller sample sizes.

An a-priori evaluation of a fitness function, without running it in the optimization loop, requires computing the ranking metric on a curated dataset.
However, each fitness function steers the search-path during the optimization.
This causes a distribution shift~\cite{lyuLearnabilityOfflineModelBased2026} between the curated dataset the metric is computed on and the topologies evaluated during the optimization.
We therefore compare two datasets, both scored with the fitness function under test and the ACOPF fitness. The \emph{ACOPF dataset} holds all topologies evaluated during 1000 generations of optimization with ACOPF in the loop.
It is built once per case and operating point (OP) and can be reused to screen any other fitness function. The \emph{own dataset} holds all topologies evaluated during the optimization with the respective fitness function itself.

\section{Case study setup}\label{sec:setup}

\begin{figure*}[t]
\centering
\includegraphics[width=\textwidth]{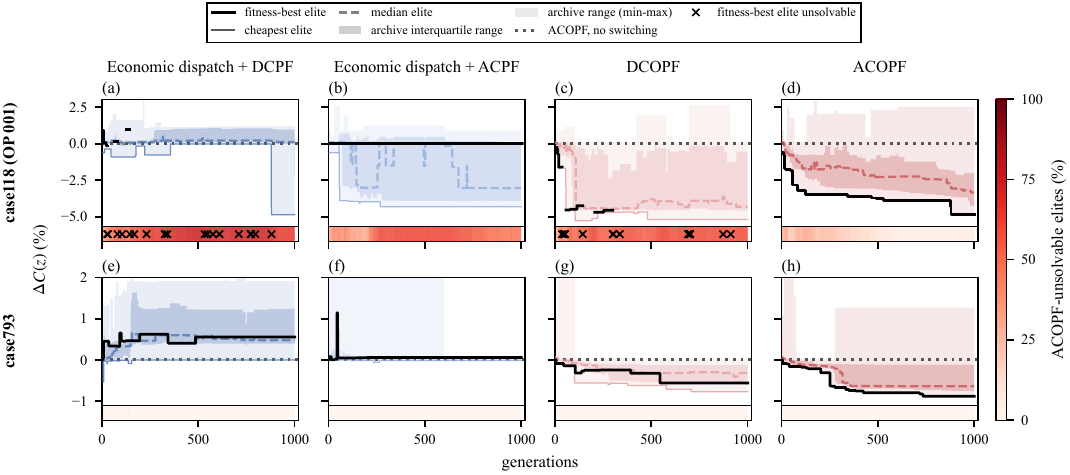}
\caption{$\Delta C(z)$~\eqref{eq:deltaC} over the generations for one run (seed 0) per fitness function, for case118 OP 001 (top, illustration only) and case793 (bottom). Lines: fitness-best elite (thick), cheapest elite (thin) and median elite (dashed) of the archive; shaded: interquartile and min--max range of the archive. The strip under each panel gives the share of ACOPF-unsolvable elites in the archive; $\times$ marks where an ACOPF-unsolvable elite becomes the fitness-best elite.}
\label{fig:convergence}
\end{figure*}

We use the ToOp default parameters: the search varies the assignment of lines to busbars, while generators and loads stay fixed to their initial busbar, i.e., one injection placement per split~\cite{westerbeckTransmissionTopologyOptimization2026}.
Topology changes are limited to a maximum of four busbar splits.
Each run has 1000 generations.
The archive has 225 cells: 0 to 4 split substations times switching distances 0 to 44.
All power flow and OPF evaluations use pandapower.
The case study runs four seeds per OP to address randomness in the action enumeration and the MAP-Elites search; the ranking results in Section~\ref{sec:ranking} use only one seed due to limited compute resources.

The case study investigates two cases from PGLib~\cite{babaeinejadsarookolaeePowerGridLibrary2021}: an IEEE 118-bus system (case118) and a 793-bus system (case793) from the GO competition.
For case118 we create 50 congested OPs by sampling the active load vector from a Dirichlet distribution and scaling it up in steps.
We keep the last load level at which the base-topology ACOPF with line limits is still feasible and at least five lines are at 100\,\% of their rating, similar to the PGLib API cases.
Case793 uses its PGLib OP, not the API variant, because the base case is already congested under economic dispatch.
The economic dispatch overloads both cases for all OPs.

\section{Results}\label{sec:results}

\subsection{Optimization performance of the fitness functions}\label{sec:performance}

\begin{figure}[]
\centering
\includegraphics[width=\columnwidth]{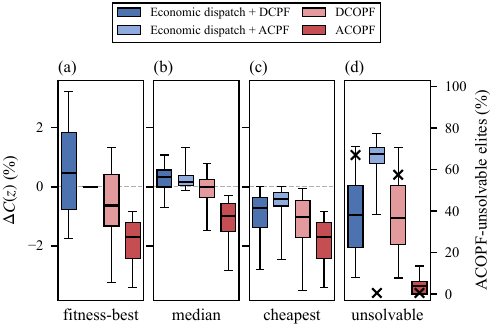}
\caption{Optimization outcome per fitness function on case118, over 50 OPs $\times$ 4 seeds: (a)--(c) $\Delta C(z)$ of the fitness-best, median and cheapest elite for the ACOPF-solvable cases. (d) shows the share of ACOPF-unsolvable elites, and the crosses mark the share of runs whose fitness-best elite is ACOPF-unsolvable. Boxes span the interquartile range, the line is the median, and the whiskers span the 5th to 95th percentile.}
\label{fig:performance}
\end{figure}

Fig.~\ref{fig:convergence} shows the redispatch cost change $\Delta C(z)$ and the ACOPF solvability during the optimization of the elites for an exemplary case118 OP (OP 001) and case793.
Fig.~\ref{fig:performance} aggregates the performance of the different fitness functions over the 50 case118 OPs in the form of $\Delta C(z)$ and the share of ACOPF-solvable topologies in the final elites.

\emph{PF-based fitness functions.}
The fitness-best elite of the PF-based fitness functions does not necessarily reduce the redispatch cost of the base topology, but often increases it, see Fig.~\ref{fig:performance}.
This points to a mismatch of the objective. The PF-based fitness functions evaluate how well a topology performs for a fixed economic dispatch, but the ACOPF cost is the minimum redispatch cost over all feasible dispatches.
Thus, with a PF-based fitness function, MAP-Elites optimizes the topology for overload reduction of a given economic dispatch instead of its redispatch cost reduction.
However, in the diverse set of elites, there are topologies that reduce the redispatch cost, as shown by the distribution of elites in case118 in Fig.~\ref{fig:convergence}, and the cheapest elite in Fig.~\ref{fig:performance}.
Nevertheless, case793 in Fig.~\ref{fig:convergence} shows that this is not guaranteed: here, no elite improves over the base topology. This supports the findings of \cite{bannmullerAddressingModelInaccuracies2025,crozierModelingGenerateAlternatives2025a}.
Looking at the ACOPF solvability of the elites, the DCPF and DCOPF fitness functions lead to similar numbers of unsolvable elites. This points to a feasibility misjudgment of the DC models, independent of the dispatch coupling.
The ACPF fitness function results in a higher number of unsolvable elites. On case118, its fitness-best elite is the base topology in 199 of 200 runs, because the search finds no topology with less AC overload than the base topology.
The optimization leads to topologies for which the ACPF converges, but not necessarily to topologies where redispatch can keep power flows below the line limits.
The ACPF feasibility estimate is therefore not necessarily stricter than the reference ACOPF fitness function; for the line limits, it is loose.
In case793, the feasibility misjudgment is less pronounced, perhaps because this OP and the larger grid are not as sensitive to busbar splits as case118.

\emph{OPF-based fitness functions.}
OPF-based fitness functions co-optimize the redispatch for each topology, therefore evaluating the topology's redispatch potential when used in the optimization loop.
This results in improved optimization performance over the PF-based ones: both DCOPF and ACOPF already reduce the ACOPF cost with their fitness-best elite in both cases.
The sets of elites of DCOPF and ACOPF also outperform those of DCPF and ACPF, especially for case793.
However, the results reveal that, for individual OPs, the cheapest elites of PF-based fitness functions can outperform OPF-based ones. Here, some ACPF elites perform better than ACOPF elites for several generations, as shown in Fig.~\ref{fig:convergence}.

\subsection{Does the ranking metric predict optimization performance?}\label{sec:ranking}

\begin{figure}[]
\centering
\includegraphics[width=0.85\columnwidth]{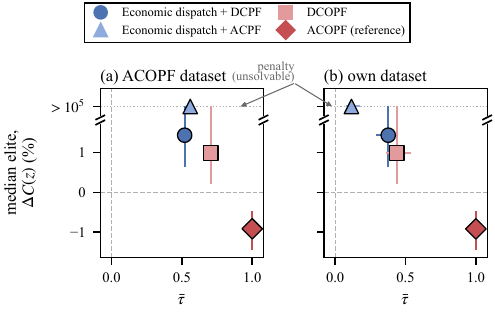}
\caption{Per-cell median Kendall $\bar{\tau}$~\eqref{eq:taucell} against the median-elite $\Delta C(z)$ on case118 for one seed (same as the datasets), with $\bar{\tau}$ computed on (a) the ACOPF dataset and (b) each fitness function's own dataset. Ranking and elite performance evaluation are based on fitness including unsolvability-penalty.}
\label{fig:tau_vs_performance}
\end{figure}

Fig.~\ref{fig:tau_vs_performance} shows the median elite performance over the per-cell median Kendall $\bar{\tau}$~\eqref{eq:taucell} of the four fitness functions on 50 OPs of case118, with $\bar{\tau}$ computed on the ACOPF dataset and on each fitness function's own dataset.
Here, unlike in Fig.~\ref{fig:performance}, the cost includes the penalty for unsolvable topologies, so that $\bar{\tau}$ also reflects the feasibility misjudgment of the fitness function.
The results show that the evaluation of the ranking metric on the ACOPF dataset does not reflect the ordering of optimization performance of the fitness functions, while the evaluation on each fitness function's own dataset does.
In particular, the $\bar{\tau}$ of the ACPF fitness function drops drastically when evaluated on its own dataset compared to the ACOPF dataset.
This indicates that the ranking performance of the ACPF fitness function is worse on the topologies it visits during the optimization.
Note that the own dataset is only available after the optimization has run and additional ACOPF runs have evaluated its topologies as the ranking reference: $\bar{\tau}$ on it diagnoses the distribution shift, but is not an a-priori prediction.
The distribution shift is also reflected in the low overlap of the topologies visited by the different optimization search paths: per OP, the own datasets share a median of 11--14\,\% of topologies with the ACOPF dataset.

\section{Conclusion}\label{sec:conclusion}

While the exact optimization performance is grid- and operating-point-dependent, OPF-based fitness functions outperform PF-based ones on the test cases.
Optimization with PF-based fitness functions depends on the diversity of the MAP-Elites archive. In the cases studied, this often leads to redispatch cost reduction, but there is no guarantee.
Co-optimizing the redispatch evaluates a topology's redispatch potential instead of its performance for a fixed economic dispatch.
This co-optimization shows to be more important than increasing the model resolution from DC to AC.
Future work should investigate real transmission systems and OPs to validate the findings, and detailed redispatch modeling.

Furthermore, the proposed ranking-based metric shows that the evaluation of fitness functions on a curated dataset of topologies can estimate the ordering of their optimization performance.
However, this requires a small distribution shift between the dataset and the topologies visited during the optimization, which can only be assessed after the optimization.
Future work should address this distribution shift, so that the proposed metric can be used to develop fitness functions that are tailored to NTO with gradient-free optimization algorithms.

\section*{Acknowledgments}
The work of P. Bannmüller and J.L. Cremer is supported through the AI-EFFECT project (Grant Agreement No.~101172952) funded under the European Union's Horizon Europe Research and Innovation program. However, views and opinions expressed are those of the author(s) only and do not necessarily reflect those of the European Union. Neither the European Union nor the granting authority can be held responsible.
Claude Opus (Anthropic) has been used to write code and rephrase parts of this paper. The authors take full responsibility for the content of this publication.

\bibliographystyle{IEEEtran}
\bibliography{bib/IEEEtran_control,bib/Bib_complete}

\end{document}

%% file: figures/surrogate_schematic.tex
\pgfplotsset{
  schematic/.style={
    scale only axis,
    width=7cm,
    height=3.3cm,
    xmin=0, xmax=10,
    ymin=-0.25, ymax=4.5,
    axis lines=left,
    axis on top,
    xtick=\empty, ytick=\empty,
    clip=false,
    xlabel={topology},
    ylabel={cost},
    label style={font=\footnotesize},
    xlabel style={yshift=2pt},
    ylabel style={yshift=-2pt},
    samples=300,
    legend style={font=\footnotesize, draw=none, fill=none,
      at={(0.5,1.03)}, anchor=south, legend columns=2,
      /tikz/every even column/.append style={column sep=6pt}},
  },
}
\colorlet{tightcolor}{red!80!black}
\colorlet{loosecolor}{blue!70!cyan}
\colorlet{goodcolor}{green!55!black}
\tikzset{
  truecost/.style={black, line width=0.8pt},
  tightsurrogate/.style={tightcolor, line width=0.8pt, dashed},
  loosesurrogate/.style={loosecolor, line width=0.8pt, dash dot},
  goodsurrogate/.style={goodcolor, line width=1.0pt, densely dotted},
  infeasibleband/.style={fill=black!8, draw=none},
  pickmarker/.style={circle, inner sep=1.5pt, draw=#1, fill=#1},
  annotation/.style={font=\scriptsize, align=center, inner sep=1pt},
}
\begin{tikzpicture}
\begin{axis}[
  schematic,
  declare function={
    edgegap(\d) = max(0, 1/\d - 1/0.6);
    barrier(\x,\l,\r) = 0.08*(edgegap(\x-\l)*edgegap(\x-\l)
      + edgegap(\r-\x)*edgegap(\r-\x));
    truecost(\x) = 2.2
      - 0.6*exp(-((\x-1.6)*(\x-1.6))/0.15)
      - 1.45*exp(-((\x-3.7)*(\x-3.7))/0.35)
      - 1.9*exp(-((\x-6.3)*(\x-6.3))/0.45)
      - 0.7*exp(-((\x-7.4)*(\x-7.4))/0.12);
    loosesurrogate(\x) = 0.04 + 0.014*(\x-9.2)*(\x-9.2);
    goodsurrogate(\x) = 3.0
      - 0.8*exp(-((\x-3.7)*(\x-3.7))/0.7)
      - 1.7*exp(-((\x-6.3)*(\x-6.3))/0.9);
  },
]
\fill[infeasibleband] (axis cs:0,-0.25) rectangle (axis cs:1.2,4.5);
\fill[infeasibleband] (axis cs:8.6,-0.25) rectangle (axis cs:10,4.5);
\node[annotation, rotate=90, text=black!55] at (axis cs:0.6,3.0) {redispatch-\\infeasible};
\node[annotation, rotate=90, text=black!55] at (axis cs:9.3,3.3) {redispatch-\\infeasible};

\addplot[truecost, domain=1.334:8.458, samples=600]
  {truecost(x) + barrier(x,1.2,8.6)};
\addlegendentry{redispatch cost}

\addplot[tightsurrogate, domain=2.359:4.743, samples=400]
  {truecost(x) + barrier(x,2.2,4.9)};
\addlegendentry{tight fitness function}

\addplot[loosesurrogate, domain=0:10] {loosesurrogate(x)};
\addlegendentry{loose fitness function}

\addplot[goodsurrogate, domain=1.530:8.271, samples=600]
  {goodsurrogate(x) + barrier(x,1.35,8.45)};
\addlegendentry{good fitness function}

\node[pickmarker=black] (trueopt) at (axis cs:6.3,0.30) {};
\node[pickmarker=goodcolor] (goodpick) at (axis cs:6.3,1.30) {};
\draw[goodcolor, thin] (goodpick) -- (trueopt);
\node[annotation] (globallabel) at (axis cs:7.75,0.55) {global\\optimum};
\draw[black, -{latex}, thin] (globallabel.west) -- (trueopt);
\node[pickmarker=tightcolor] (tightpick) at (axis cs:3.7,0.75) {};
\node[annotation, text=tightcolor, anchor=south] at (axis cs:3.55,3.0)
  {local\\optimum};
\draw[tightcolor, -{latex}, thin] (axis cs:3.55,3.0) -- (tightpick);
\node[pickmarker=loosecolor] (loosepick) at (axis cs:9.2,0.04) {};
\node[annotation, text=loosecolor, anchor=south] at (axis cs:9.3,1.0)
  {infeasible\\pick};
\draw[loosecolor, -{latex}, thin] (axis cs:9.3,1.0) -- (loosepick);
\end{axis}
\end{tikzpicture}

%% file: figures/mapelites_schematic.tex
\colorlet{mefitlow}{black!12}
\colorlet{mefitmid}{black!32}
\colorlet{mefithigh}{black!58}
\colorlet{meaccent}{red!80!black}
\tikzset{
  mecell/.style={draw=black!45, line width=0.3pt},
  mefit1/.style={fill=mefitlow},
  mefit2/.style={fill=mefitmid},
  mefit3/.style={fill=mefithigh},
  mestep/.style={draw, rounded corners=2pt, line width=0.5pt, align=center,
    font=\scriptsize, minimum height=0.5cm, minimum width=1.45cm,
    inner sep=2pt},
  meflow/.style={-{Latex[length=1.6mm]}, line width=0.5pt},
  methin/.style={-{Latex[length=1.3mm]}, line width=0.35pt},
  metoken/.style={circle, draw, line width=0.4pt, fill=#1, inner sep=0pt,
    minimum size=0.34cm, font=\scriptsize},
  meswatch/.style={draw=black!60, line width=0.3pt, minimum size=0.2cm,
    inner sep=0pt, fill=#1},
  meparent/.style={circle, draw=black, fill=white, line width=0.4pt,
    inner sep=0pt, minimum size=0.11cm},
  melabel/.style={font=\scriptsize, inner sep=1pt},
  meaxis/.style={-{Latex[length=1.4mm]}, line width=0.4pt},
}
\def\meoccupancy{%
  {1,0,2,0,0,3},%
  {2,3,2,1,3,0},%
  {1,2,3,0,1,0},%
  {0,1,2,1,0,0}}%
\def\mecellsize{0.42}
\begin{tikzpicture}[x=1cm, y=1cm]
\foreach \row [count=\rowindex from 0] in \meoccupancy {
  \foreach \level [count=\columnindex from 0] in \row {
    \ifnum\level=0
      \draw[mecell, fill=white]
        (\columnindex*\mecellsize, \rowindex*\mecellsize)
        rectangle ++(\mecellsize, \mecellsize);
    \else
      \draw[mecell, mefit\level]
        (\columnindex*\mecellsize, \rowindex*\mecellsize)
        rectangle ++(\mecellsize, \mecellsize);
    \fi
  }
}
\coordinate (cellOne) at (4.5*\mecellsize, 1.5*\mecellsize);
\coordinate (parentOne) at (1.5*\mecellsize, 3.5*\mecellsize);
\coordinate (parentTwo) at (2.5*\mecellsize, 2.5*\mecellsize);

\draw[meaccent, line width=1.1pt] (4*\mecellsize, 1*\mecellsize)
  rectangle ++(\mecellsize, \mecellsize);
\node[melabel, text=white] at (cellOne) {$z_1$};

\draw[meaxis] (-0.1, 0) -- (-0.1, 4*\mecellsize + 0.15);
\node[font=\footnotesize, rotate=90, align=center, anchor=south]
  at (-0.12, 2*\mecellsize) {\# split\\[-2pt]substations};
\draw[meaxis] (0, -0.1) -- (6*\mecellsize + 0.15, -0.1);
\node[font=\footnotesize, anchor=north] at (3*\mecellsize, -0.12)
  {switching distance};

\node[mestep] (create) at (4.05, 3.0) {create\\[-1pt]offspring};
\node[mestep] (fitness) at (6.45, 3.0) {fitness\\[-1pt]$F_s(z)$};
\node[mestep] (mapping) at (6.45, 0.95) {descriptor\\[-1pt]mapping};

\draw[line width=0.5pt, rounded corners=2pt] (3.25, 0.35) rectangle (5.1, 1.55);
\node[font=\scriptsize] at (4.175, 1.39) {competition};
\node[metoken=mefithigh, text=white] (compOne) at (3.5, 1.0) {$z_1$};
\node[melabel] at (3.88, 1.0) {$>$};
\node[meswatch=mefitlow] at (4.25, 1.0) {};
\node[melabel] at (4.75, 1.0) {$\checkmark$};
\node[metoken=mefitlow] (compTwo) at (3.5, 0.6) {$z_2$};
\node[melabel] at (3.88, 0.6) {$\le$};
\node[meswatch=mefithigh] at (4.25, 0.6) {};
\node[melabel] at (4.75, 0.6) {$\times$};

\node[meparent] at (parentOne) {};
\node[meparent] at (parentTwo) {};
\draw[methin] (parentOne) to[out=80, in=180] ($(create.west)+(0,0.08)$);
\draw[methin] (parentTwo) to[out=60, in=180] ($(create.west)+(0,-0.08)$);
\node[melabel, align=center, anchor=south] at (2.75, 2.0)
  {parent\\[-1pt]selection};

\draw[meflow] (create) -- (fitness);
\node[metoken=white] at (5.06, 2.62) {$z_1$};
\node[metoken=white] at (5.44, 2.62) {$z_2$};

\draw[meflow] (fitness) -- (mapping);
\node[metoken=mefithigh, text=white] at (5.75, 1.95) {$z_1$};
\node[metoken=mefitlow] at (6.13, 1.95) {$z_2$};

\draw[meflow] (mapping) -- (5.1, 0.95);

\draw[methin] (compOne.west) -- (5*\mecellsize, 1.5*\mecellsize);
\draw[methin, dashed] (compTwo.west) -- (6*\mecellsize + 0.12, 0.5*\mecellsize);
\node[font=\footnotesize, text=meaccent] at (6*\mecellsize + 0.3, 0.12)
  {$\times$};

\node[melabel, anchor=east] at (4.5, -0.32) {fitness: low};
\node[meswatch=mefitlow] at (4.65, -0.32) {};
\node[meswatch=mefitmid] at (4.9, -0.32) {};
\node[meswatch=mefithigh] at (5.15, -0.32) {};
\node[melabel, anchor=west] at (5.3, -0.32) {high};
\end{tikzpicture}